# Detection of the Group of Traffic Signs with Central Slice Theorem

Koba Natroshvili

*Abstract*— Our sensor system consists of a combination of Photonic Mixer Device - PMD and Mono optical cameras. Some traffic signs have stripes at 45°. These traffic signs cancel different restrictions on the road. We detect this class of signs with Radon transformation. Here the Radon transformation is calculated using Central Slice Theorem. We approximate the slice of spectrum by the Discrete Cosine Transformation (DCT).

## I. INTRODUCTION

Plenty of publications are available in the field of traffic sign detection. Traffic sign detection is a vital part of the Advanced Driver Assistance Systems (ADAS). It helps the driver to have the actual tempo limit information. One class of traffic signs are 'end of restriction' signs. Normally traffic sign detections are divided in two steps. At first the shape is detected. Later the detected shapes are classified. Shapes generally can be circular, rectangular and triangular. The traffic sign shapes used in Europe for speed information are circular. Classification methods like Ada-Boost algorithm, Neural Networks, Support Vector Machines etc. could be used as a supervised learning tool. We use Support Vector Machines.

In the beginning we describe our camera systems consisting of time of flight principle - PMD and CMOS cameras. At next we introduce Radon transformation and the Central Slice Theorem.

## II. SENSOR SYSTEM

Our observation system includes PMD and CMOS cameras. Both of them are mounded on the body of the vehicle (Fig. 1). PMD camera combines optoelectronic system, delivering the amplitude and range images simultaneously at frame rates of 80Hz. It uses modulated infrared illumination at wavelength 850nm and modulated frequency 1,...,16MHz.

In combination with PMD we use a CMOS camera. PMD camera delivers object coordinates in the world frame. The PMD camera has small (relative to CMOS camera) resolution - 64x16 and small opening angles 55° in horizontal and 18° in vertical direction (this parameters are specific to our optical setup). The combination of PMD and CMOS cameras improves the precision of object detection and classification. The methods of detection and classification of traffic signs using optical images is given in [11], [12]. More advanced work on the supplementary traffic symbol detection could be found in [13].

## III. PRINCIPLE OF DETECTION

PMD camera delivers the 3D object list. From the object list we filter the objects. Normally traffic signs appear in the particular areas of the street. Therefore, the objects are filtered in dependence of size and position. Because of the relatively small resolution of the PMD camera, in the distance starting from 20m the mistake even for one pixel could be vital in decision. Hence, we reject only the objects that are not definitely a traffic sign. 3D objects list is projected on the 2D image by perspective mapping. Later classification is done in the 2D image. In order to limit the region of a traffic sign better the circular Hough transformation is applied.

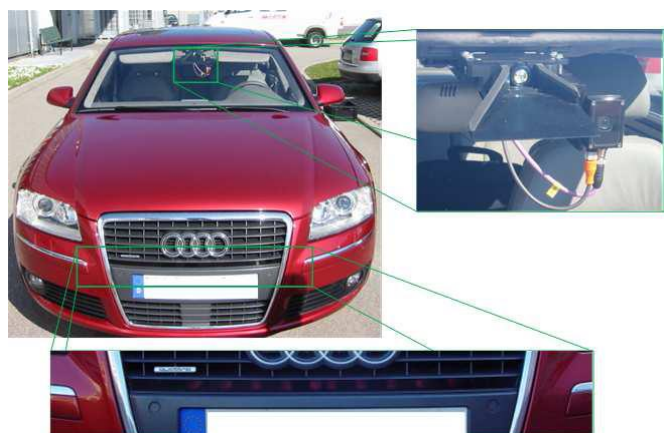

Fig. 1, 2D-3D (CMOS - PMD) Camera system mounted on the car (upper left image), close look of the camera system (upper right) illumination source of the PMD camera

It is well know that the Hough transformation can be time consuming because of the accumulator principle. Here we have used an optimized version of Circular Hough transformation, minimizing the calculation time, similar to idea developed in [10]. Later the traffic signs in the circular area are classified with the statistical supervised learning method Support Vector Machine (SVM). It finds the global minimum and solves the structural risk minimization problem. Therefore, only training examples which are close to the decision boundary become important. In the Neural Networks contrary, the problem is solved by minimizing the empirical error of the training data by finding a local minimum. SVM give very good detection performance, but calculation time in the testing phase could become large, critical for the real time applications. Therefore, the methods reducing the classification speed are very relevant to our application. We have already done a work to reduce the calculation time for binary SVM. We are able to reduce the number of support Vectors by the factor of 50-100. The details are published in [1].

In this publication we detect the class of traffic signs containing 'end of restriction' signs. It is done before the classification with SVM. In the positive detection case the

Koba Natroshvili is with Harman/Becker Automotive Systems, Becker-Göring-Str. 16, Karlsbad, Germany, koba.natroshvili@Harman.com
Submitted for IEEE Intelligent Vehicle 2010 symposium

classification is done only inside this class boosting the overall classification time.

## IV. RADON TRANSFORMATION

The Radon transformation was first introduced in [9]. A very nice description is given in [7], which we follow here as an introduction. The Radon transformation converts lines from the Cartesian coordinates into points in the polar domain. Suppose that we have a 2D function $u(x_1, x_2)$ and we are looking for the parallel projection of $u(x_1, x_2)$ along the direction $T$ as seen in Fig. 2 The parallel projection of a 2D signal can be understood as a convolution with a Dirac line perpendicular to the line of projection.

Now we convert the Dirac line to polar coordinates:

$$\delta(\mathbf{g} \cdot \mathbf{x}) = \delta(R) \cdot 1(T) \quad (1)$$

$\mathbf{g}$ is a unit vector normal to the Dirac line. The 2D signal $u(x_1, x_2)$ in polar coordinates is given as: $u_\varphi(R,T) = u(x_1, x_2)$. The result of the convolution of the Dirac line with $u_\varphi(R,T)$ is a parallel projection, and we abbreviate it as $u_p(R,T)$:

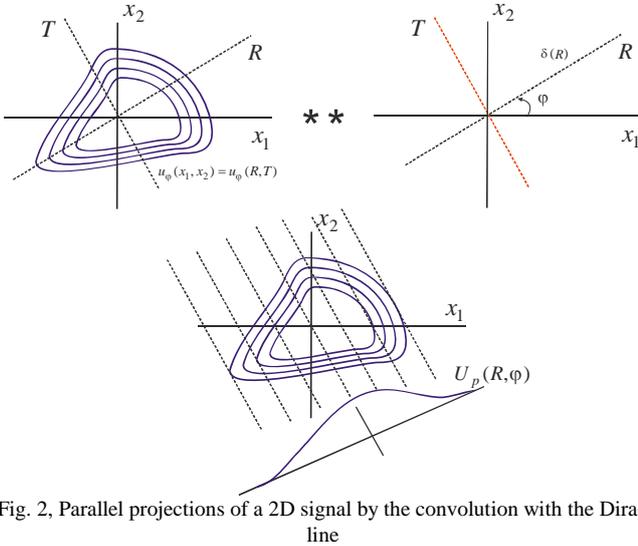

Fig. 2, Parallel projections of a 2D signal by the convolution with the Dirac line

Using (1) we express the parallel projections using the line integral in the polar coordinates:

$$u_p(R,\varphi) = \int_{-\infty}^{\infty} u_\varphi(R,T) \cdot dT = u_\varphi(R,T) ** \delta(R) \quad (2)$$

The integral given in (2) expresses the Radon transformation. The Radon transformation calculates the integrals over the lines and then gives the result in polar coordinates. It is widely used in the fields where projections in the different directions are used to reconstruct the objects in higher dimensional space (like in computer tomography).

## V. CENTRAL SLICE THEOREM

Very often, the Radon transformation is associated with the Central Slice theorem. The reason is given below. Generally for any 1D signal, we can write:

$$\int_{-\infty}^{\infty} u(t) dt = U(0) \quad (3)$$

This property comes directly from the Fourier integral after inserting the value $f = 0$. The result of (3) can be generalized for multidimensional signals:

$$\int_{-\infty}^{\infty} u(\mathbf{x}) d\mathbf{x} = U(\mathbf{0}) \quad (4)$$

where $\mathbf{0} = (0,0,...,0)^T$ is a null vector.

(4) was obtained after the integration of Fourier integral for all variables. Now it is interesting to find out what kind of result we will get if we perform the integration with only some variables and then do projection across the rest of the variables. We consider this task for 2D function. We use $R, T$ polar coordinates as shown in Fig. 2. We abbreviate the frequency axes as $f_R, f_T$. They are oriented in the same way as $R, T$ axes. We calculate the Fourier spectrum as:

$$\begin{aligned} u(x_1, x_2) &=: u_\varphi(R,T) \circ\!\!\leftrightarrow\!\!\bullet \\ U_\varphi(f_R, f_T) &:= U(f_1, f_2) \end{aligned} \quad (5)$$

which is equivalent to:

$$U_\varphi(f_R, f_T) = \int_{-\infty}^{\infty}\int_{-\infty}^{\infty} u_\varphi(R,T) e^{-j2\pi(R \cdot f_R + T \cdot f_T)} dR dT \quad (6)$$

By setting $f_T = 0$, we get

$$\begin{aligned} U_p(f_R, \varphi) &= \int_{-\infty}^{\infty} \left[ \int_{-\infty}^{\infty} u_\varphi(R,T) dT \right] \cdot e^{-j2\pi R \cdot f_R} dR \\ &= \int_{-\infty}^{\infty} u_p(R,\varphi) \cdot e^{-j2\pi R \cdot f_R} dR \end{aligned} \quad (7)$$

$U_p(f_R, \varphi)$ is the slice of 2D spectrum at the origin and taken in $\varphi$ direction. ($u_p(R,\varphi)$ is the projection of the 2D function $u(x_1, x_2)$ in direction $T$ evaluated across the $R$ axes).

The equation above represents the Central Slice theorem. We write it in another way:

$$\begin{aligned} \int_{-\infty}^{\infty} u_\varphi(R,T) dT &= u_p(R,\varphi) \circ\!\!\leftrightarrow\!\!\bullet \\ U_p(f_R, \varphi) &= U_\varphi(f_R, f_T = 0) \end{aligned} \quad (8)$$

The Central Slice theorem can be formulated in the following way: the spectrum of the projected signal can be calculated by taking the slice of the original signal spectrum; the slice should be taken at the origin and at the perpendicular to the projection direction. There are different interpretations of the Central Slice theorem for example see [2], [3], [4], [5], [6].

## VI. APPROXIMATION OF FOURIER SPECTRUM WITH DCT

As we have mentioned above here we calculate the projections using the Radon transformation. The Radon transformation is calculated using Central Slice Theorem. Normally the Fourier spectrum is estimated by the Fast Fourier Transformation (FFT), which is implemented in most of the software libraries. In our case in the classification phase with SVM we use DCT as the feature attributes. Therefore it is profitable to directly use DCT instead of calculation additionally FFT. In order to save the calculation time we approximate FFT with DCT. In DCT the signal is represented as the sum of cosine functions. Between FFT and DCT there is a similarity, but differences as well. The major difference is using only real values in the integration in DCT opposite to FFT where in complex numbers appear in the summation.

## VII. EVALUATION OF THE DESCRIBED APPROACH ON IDEAL IMAGES

In the beginning, we have made some simulation on ideal traffic sign images. We used the high resolution images (corresponding to the close distance) and they are free of noise or any blurring effect. At first we generated the 2D Fourier spectrum of different traffic sign images. In Fig. 3 we have the absolute value of the spectrum of two different traffic signs. In A) it is for 'end of all restrictions' traffic sign. We can see that the most important part of the spectrum is across the diagonal line at 45°. It is perpendicular to the 5 lines in the image which are at 45°. Here it should be mentioned that the origin of the traffic sign image is at the top left (normal image visualization convention). Therefore the image should be rotated before the comparison.

Very different results we obtain in case of the second traffic sign ('no passing') which is shown in B). Here spectrum does not have any diagonal form. This is first demonstration of the Central Slice Theorem.

In order to apply the Central Slice Theorem we cut the slice of spectrum at 45° and apply the inverse Furrier transformation. The results are available in Fig. 4. To make a comparison we made the projections in three different directions in horizontal (black), vertical (yellow) and along 45° corresponding to the 5 lines (blue). We observe that on the blue plot we have 5 peaks corresponding to the lines on the traffic sign.

As we have already mentioned we use DCT as the features for the training with SVM. Therefore in order to avoid extra transformation and save calculation time we have tried to substitute FFT with DCT.

Results of approximation of projections with DCT we have visualized in Fig. 5. In reality lines in the end of restriction traffic signs are dark on the white background. Therefore instead of maximums we get minimums at the line locations. In the middle plot we have again projections in three different directions. Blue corresponds to 45°, which is interesting for us. We see that instead of 5 minimums we have only 3. This error comes from the approximation when we replace FFT with DCT.

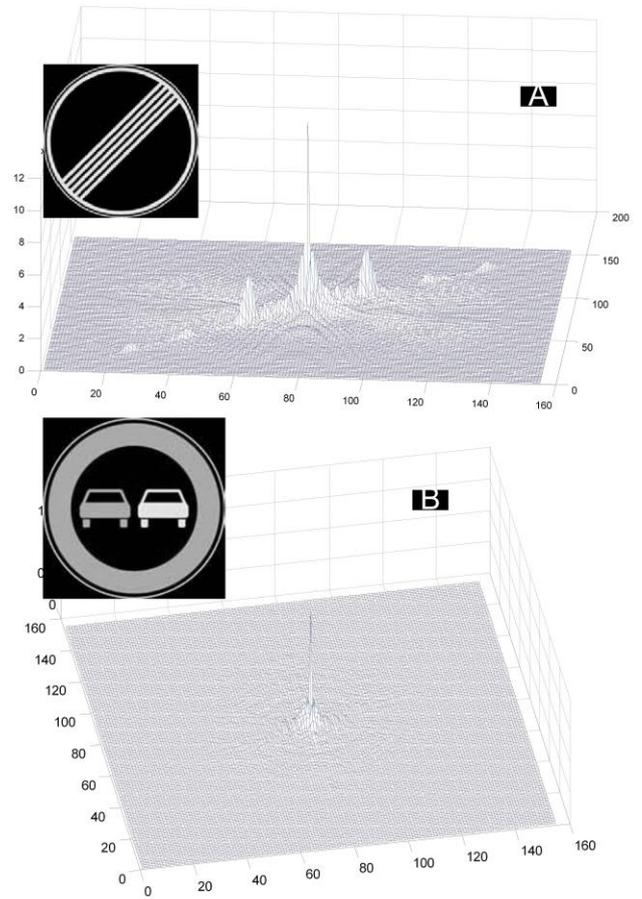

Fig. 3, A) absolute value of 2D spectrum of 'end of all restrictions' traffic sign, B) absolute value of 2D spectrum of 'no passing' sign

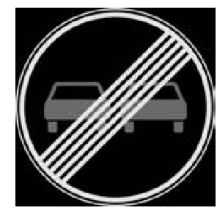

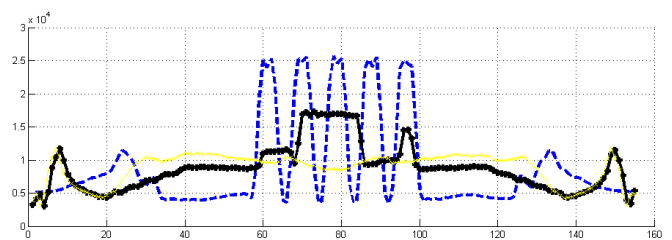

Fig. 4, Results of projections across horizontal line (black), vertical (yellow) ,45° (blue)

Lower plot of Fig. 5 is again the signal of 45° projection, but now it is normalized and filtered. Normally direct back projection gives blurred results (see [8]). It is advisable to us ramp $|f|$ type of filter. It attenuates linearly the low frequency magnitudes and amplifies the high ones. It appears as a weighting factor when calculating the Jacobean of Cartesian to polar coordinate transformation.

We search for the characteristic minimum at the right part of the signal and detect the lines at 45°.

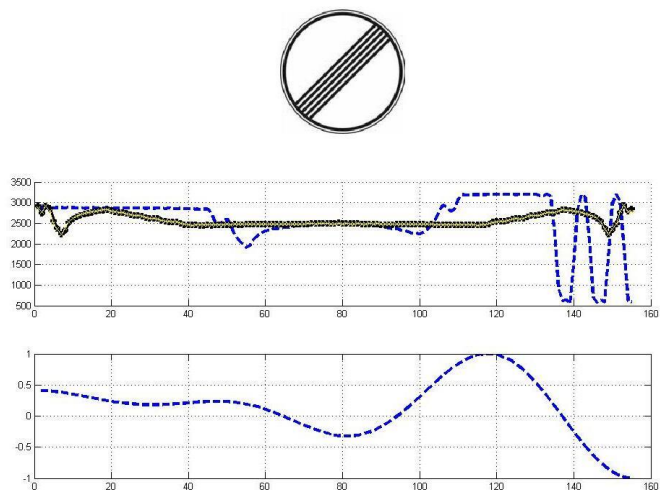

Fig. 5, Results of projection with DCT blue is 45° degree in the upper part; the same projection but filtered and normalized with Ramp filter

## VIII. EVALUATION OF THE DESCRIBED APPROACH ON REAL IMAGES

In order to test our algorithm in realistic scenarios we have evaluated it on the traffic sign images taken by a camera on the streets. We have concentrated on the following traffic signs: 'maximum speed limits: 30, 60, 100', speed prohibitions, 'end of speed limit 50, 60, 80' and 'end of all restrictions'.

We have tested our algorithm on the traffic sign images with sizes of 20x20 pixels. It corresponds to a distance of about 25m. Because of the different weather and day conditions and the small size of the images, they look very blurred. Besides, 5 lines couldn't be separated any more and they are blurred to one single line.

The number of examples and results of the correct detections are given in Table 1. In some cases (for example for 'end of all restrictions') the results of detection is not so good. It could be explained by the bad quality of images. Some images are so blurred that it is difficult to identify them even with human eye.

Besides, we have tested our algorithm with 7685 of no traffic sign images. We have got 0.6% of false positives which is a really good result.

## RESULTS

'End of restrictions' traffic signs have characteristic five stripes at 45°. We made a projection of a 2D image along these lines and identify these stripes. Projection was made by Radon transformation. It was calculated using Central Slice Theorem. In the calculations FFT was substituted by DCT, which is an approximation. In this way we detect one class of traffic signs containing end of different restriction signs, which saves the time of classification. Detection rate was more that 86% on the average, with 0.6% false positives.

Table 1, Parameters of the spaceborne bistatic simulation

| Traffic sign class label | Detection rate as positive | Number of examples |
|---|---|---|
| Speed limit 30 | 6% | 296 |
| Speed limit 60 | 0 | 257 |
| Speed limit 100 | 0 | 616 |
| End of all restriction | 82% | 220 |
| End of 50 | 96% | 29 |
| End of 60 | 85% | 54 |
| End of 80 | 87% | 62 |

## FUTURE PERSPECTIVES

It would be interesting to consider the projections in different directions as well. Combinations of these projections could give the better results. In the training process of SVM until now we have used grey values, DCT and Quadruple Haar Wavelet Transformation attributes. The results of the projections could be also considered as the candidates in the supervised training process.

REFERENCES

[1] K. Natroshvili, M. Schmid, M. Stephan, A. Stiegler, T. Schamm, Real Time Pedestrian Detection by Fusing PMD and CMOS Cameras, IV2008, Eindhoven, Holland
[2] K. Natroshvili, 'Bistatic Processing – Analysis and Verification', Doctoral Thesis, ZESS (Center for Sensorsystems), University of Siegen Electronic Library, 2007
[3] K. Natroshvili, O. Loffeld, H. Nies, Focusing of arbitrary bistatic SAR configurations, Proc. EUSAR 2006
[4] A. M. Ortiz, O. Loffeld, S. Knedlik, H. Nies, K. Natroshvili,, Comparison of Doppler centroid estimators in bistatic airborne SAR, Proceedings. IEEE International Geoscience and Remote Sensing Symposium, 2005
[5] U. Gebhardt, O. Loffeld, H. Nies, K. Natroshvili, S. Knedlik, Bistatic space borne/airborne experiment: Geometrical modeling and simulation, IEEE International Symposium on Geoscience and Remote Sensing, 2006
[6] Holger Nies, Otmar Loffeld, Koba Natroshvili, A Ortiz, J Ender, A solution for bistatic motion compensation, IEEE International Symposium on Geoscience and Remote Sensing, 2006
[7] R. Bamler, ‚Mehrdimensionale lineare Systeme - Fourier Transformation und δ Functionen', *Springer Verlag, 1989,* ISBN 3-540-51069-9
[8] R. Gonzalez, 'Digital Image Processing', 2006, Pearson Education
[9] J. Radon, ‚Über die Bestimmung von Functionen durch ihre Integralwerte Läangs gewisser Mannigfaltigkeiten, Berichte über die Verhandlungen der Königlich Sächsischen Gesellschaft der Wissenschaften – Mathem. Physik. Klasse 69, 262-227, 1917
[10] Barnes, N. and Zelinsky, A. Real-time radial symmetry for speed sign detection, IEEE Intelligent Vehicle Symposium, 2004
[11] D. Nienhüser, T. Gumpp, J. M. Zöllner, R. Dillmann, Recognition and Attribution of Variable Message Signs and Lanes, Intelligent Vehicles Symposium 2008, Einhoven, Holland
[12] D. Nienhüser, M. Ziegenmeyer, T. Gumpp, K.-U. Scholl, J.M. Zöllner, R. Dillmann, Kamera-basierte Erkennung von Geschwindigkeitsbeschränkungen auf deutschen Straßen, Autonome Mobile Systeme, Springer Verlag, 2007, pages 212-218

[13] D. Nienhüser, T. Gumpp, J. M. Zölner, K. Natroshvili, Fast and Reliable Recognition of Supplementary Traffic Signs, Intelligent Vehicles Symposium, 2010, San Diego, US